\documentclass[11pt]{article}
\usepackage{amsmath}
\usepackage{multirow}
\usepackage{amssymb}
\usepackage{booktabs}
\usepackage{tablefootnote}
\PassOptionsToPackage{hyperfootnotes=false}{hyperref}
\usepackage[final]{acl}

\usepackage{times}
\usepackage{latexsym}

\usepackage[T1]{fontenc}

\usepackage[utf8]{inputenc}

\usepackage{microtype}

\usepackage{inconsolata}

\usepackage{graphicx}

\title{Agora: Enhancing LLM Agent Reasoning Via Auction-Based Task Allocation}

\author{
  Kaiji Zhou, Ale\v{s} Leonardis, \and Yue Feng\thanks{Corresponding author.} \\
  School of Computer Science, University of Birmingham \\
  Birmingham, United Kingdom \\
  \texttt{kxz571@student.bham.ac.uk} \\
  \texttt{\{a.leonardis,y.feng.6\}@bham.ac.uk}
}

\begin{document}
\maketitle
\begin{abstract}
Enhancing the reasoning capabilities of large language model (LLM) agents requires effective orchestration of diverse expert models and tools. However, existing frameworks typically call  APIs, based on coarse-grained matching between tasks and the functions of expert models or tools, while overlooking critical factors such as performance variability and cost efficiency among functionally similar alternatives.
To address this, we propose Agora, a framework that uses a confidence-calibrated auction to dynamically allocate tasks to expert models and tools.
By treating reasoning steps as tradeable items, Agora bases allocation on calibrated competence rather than raw confidence.
Across five main benchmarks, Agora improves or remains competitive with single-model, routing, and cascade baselines under matched candidate pools.\footnote{Code: \url{https://github.com/KAIJUJO/Agora}}
\end{abstract}

\section{Introduction}
\label{sec:intro}

Advancing the reasoning capabilities of Large Language Models (LLMs) requires moving beyond monolithic execution. 
While strategies like Chain-of-Thought (CoT) \citep{wei2023chain} provide a structural blueprint by decomposing queries into atomic steps, executing these complex chains often exceeds the reliable scope of any single generalist model.
A critical bottleneck restricts current reasoning systems: \emph{the mismatch between task difficulty and model capability}.
Static routing ignores query-dependent differences in model suitability \citep{nguyen2025metallmhighperformantcostefficientdynamic}.
These limitations motivate \emph{dynamic competence discovery}: routing each reasoning step to a suitable agent.

However, implementing this fine-grained orchestration faces two hurdles: \textit{structural alignment} and \textit{trustworthy valuation}. First, coarse-grained routing (at the query level) fails to exploit the sub-task structures generated by planners. Second, and more critically, establishing a reliable auction is plagued by \textit{overconfidence} \citep{Huang_2025}. Agents often hallucinate certainty, claiming high confidence on incorrect answers. Without a reliable measure of an agent's \textit{true} probability of success, dynamic allocation risks assigning critical logic nodes to overconfident but incompetent agents, causing the reasoning chain to collapse.

To address these challenges, we propose \emph{Agora}, a framework that reformulates task allocation as a confidence-calibrated \emph{auction}.
Specifically, Agora operates in two phases: a \emph{Planner} decomposes the query into atomic units, and an \emph{Auction} treats these units as tradeable items.
Agents compete to solve them by submitting ``bids'' derived from their execution cost and \emph{calibrated confidence}.
Crucially, by employing a hierarchical calibration strategy—combining static baselines with online adaptation—Agora filters out hallucinated certainty.
This lets routing adapt to calibrated, step-specific estimates of agent reliability.

In summary, our contributions are as follows:

\noindent $\bullet$ \emph{Auction-Based Reasoning Framework:} We propose Agora, a framework that leverages an auction mechanism to dynamically route reasoning steps, enabling specialized agents to collaborate efficiently on complex tasks.

\noindent $\bullet$ \emph{Competence-Driven Calibration:} We introduce a strategy combining embedding-based binning with online refinement to standardize confidence estimation, ensuring that allocation is based on reliability rather than hallucinated certainty.

\noindent $\bullet$ \emph{Empirical Validation:} Across five main benchmarks, Agora improves or remains competitive under matched candidate pools while making the cost--quality trade-off directly tunable.

\begin{figure*}[t]
  \centering
  \includegraphics[width=\textwidth]{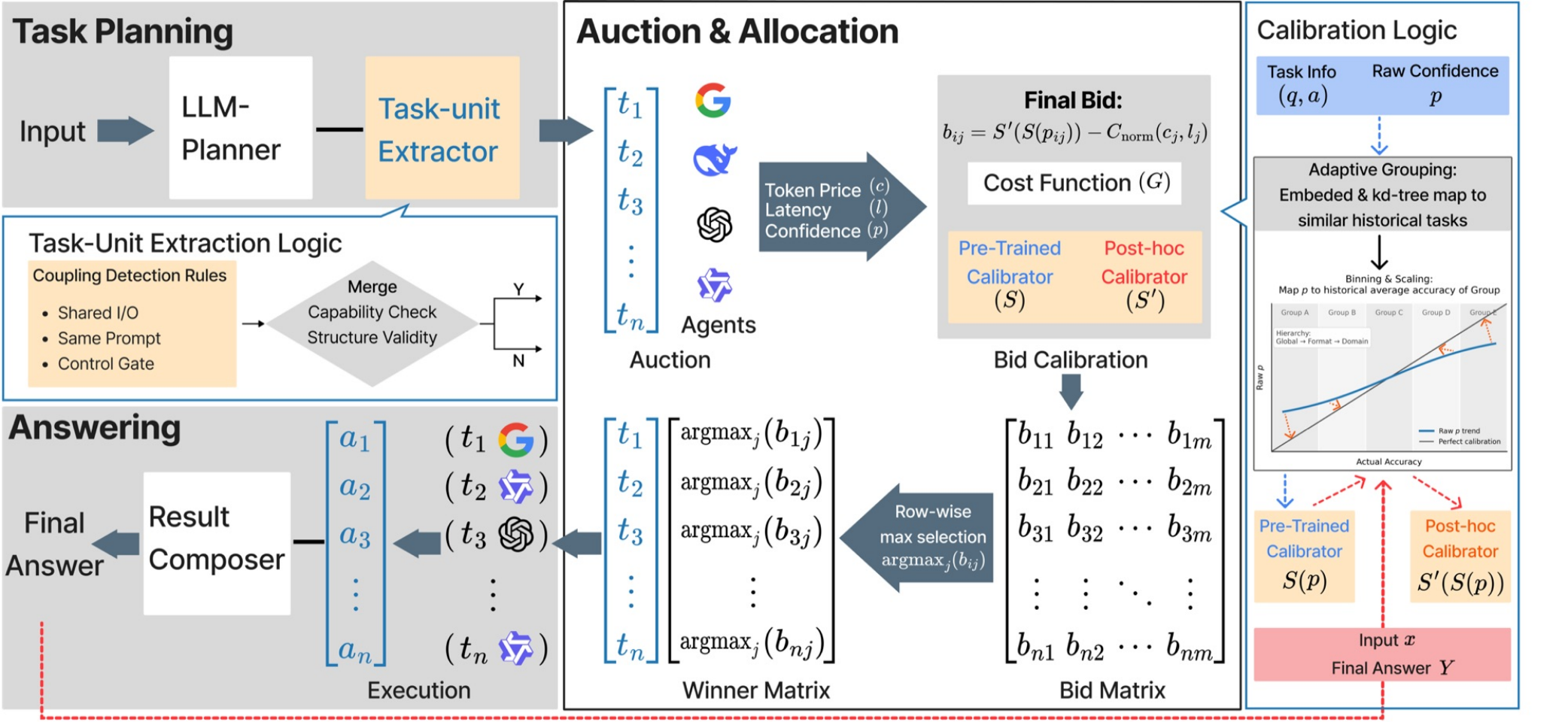}
  \caption{
Overview of Agora's \emph{auction-based reasoning framework}.
  Given a complex query $x$, a planner decomposes it into a graph of dependencies, which are grouped into \emph{task units}.
  Each unit is dynamically allocated to the highest-scoring agent via a \emph{confidence-calibrated auction} that balances calibrated competence against execution cost.
  Finally, the unit outputs are synthesized into the final answer $Y$.
  }
  \label{fig:method_overview}
\end{figure*}

\section{Related Work}
\label{sec:related_work}

\subsection{Model Selection and Specialized Reasoning}
Enhancing reasoning capabilities relies on coordinating diverse expertise. 
The prohibitive cost of generalist models has spurred research on \emph{dynamic model selection} to access specialized capabilities efficiently.
Early heuristic \emph{cascades}, such as FrugalGPT, sequentially query models to cut costs but suffer from serial latency \citep{chen2023frugalgptuselargelanguage}. 
To address this, recent work employs learned \emph{routers}. Supervised classifiers like HybridLLM \citep{ding2024hybridllmcostefficientqualityaware} and AdaptiveLLM \citep{cheng2025adaptivellm} optimize accuracy under budgets.
Advanced approaches utilize contextual bandits (RouteLLM \citep{ong2025routellmlearningroutellms}) or structural modeling (MetaLLM \citep{nguyen2025metallmhighperformantcostefficientdynamic}) to adapt to query complexity. 
Building on this, TensorOpera Router achieves significant throughput gains by projecting queries into a learned latent space to map inputs to experts \citep{stripelis2024tensoroperaroutermultimodelrouter}.

Parallel research pushes towards finer granularities. Techniques like Confidence-Token routing \citep{chuang2025learningroutellmsconfidence} and Mixture-of-Depths (MoD) \citep{raposo2024mixtureofdepthsdynamicallyallocatingcompute} optimize compute at the \emph{token} level by skipping unnecessary operations. 

Parallel to system-level routing, architectural \emph{Mixture-of-Experts (MoE)} methods have gained prominence for scaling model capacity. Architectures like Switch Transformers \citep{fedus2022switchtransformersscalingtrillion} and Mixtral \citep{jiang2024mixtralexperts} route tokens to specific internal expert layers to optimize computational efficiency. Recent innovations like DeepSeek-MoE \citep{dai2024deepseekmoeultimateexpertspecialization} further refine this by employing fine-grained expert segmentation. 
However, while these methods share the philosophy of specialization, they require end-to-end pre-training. 
In contrast, our framework focuses on \emph{auction-based orchestration}, enabling \emph{diverse agents} (e.g., proprietary planners combined with open-weight executors) to collaborate on reasoning tasks without architectural modifications.

Crucially, existing systems operate at extremes: either routing the \emph{entire query} (coarse-grained) or individual \emph{tokens} (fine-grained). 
They lack the semantic granularity to decompose complex problems into distinct \emph{sub-tasks}. 
Our framework addresses this gap by auctioning \emph{task units} using calibrated competence and execution cost.

\subsection{Calibration for Reliable Valuation}
In an auction-based system, reliable routing hinges on accurate self-valuation. 
Early ``Ask-for-calibration'' methods prompted models to verbalize certainty \citep{tian2023justaskcalibrationstrategies}, and follow-up studies refined this via prompting strategies and aggregation \citep{xiong2024llmsexpressuncertaintyempirical,yang2024verbalizedconfidencescoresllms}. 
Post-hoc methods like QA-Calibration improved reliability via embedding-based grouping \citep{manggala2025qacalibrationlanguagemodelconfidence}, yet they typically rely on static calibration sets.

Recent advancements leverage continuous semantic spaces for greater precision. Kernel Language Entropy (KLE) calculates uncertainty via semantic similarity kernels \citep{nikitin2024kernellanguageentropyfinegrained}, while Semantic Nearest Neighbor Entropy (SNNE) aggregates pairwise similarities to handle long-form generation \citep{nguyen2025semanticentropyboostingllm}.
These methods provide complementary semantic uncertainty estimates. Agora instead applies static and online calibration to self-reported confidence before allocation.

\subsection{Mechanism-Driven Reasoning}
Before modern LLM agents, task-oriented dialogue systems tracked structured states \citep{feng-etal-2021-sequence}; Chain-of-Thought \citep{wei2023chain} and Least-to-Most prompting \citep{zhou2023leasttomostpromptingenablescomplex} later made reasoning decomposition explicit.
HuggingGPT \citep{shen2023hugginggptsolvingaitasks}, Chameleon \citep{lu2023chameleonplugandplaycompositionalreasoning}, and Gradientsys \citep{song2025gradientsysmultiagentllmscheduler} dispatch sub-tasks to specialists; Med-VRAgent combines visual guidance, self-reward, and tree search for medical reasoning \citep{guo-etal-2025-med}.
However, they rarely optimize competence against cost.

To ground collaborative reasoning in economic principles, we draw on \emph{mechanism design}.
D{\"u}tting et al.\ study incentive properties for token-level LLM aggregation under explicit preference and payment rules \citep{duetting2024mechanismdesignlargelanguage}.
Inspired by this perspective, Agora treats sub-tasks as auctionable items and allocates them by balancing calibrated competence against execution cost.

Multi-agent debate can improve reasoning and factuality through social interaction \citep{du2023improvingfactualityreasoninglanguage}. However, game-theoretic evaluations reveal substantial limits in LLM rationality \citep{fan2023largelanguagemodelsserve}. This motivates calibrating self-reported confidence before allocation.

\section{Methodology}
\label{sec:method}

Our framework follows a closed-loop pipeline: \emph{Planning} $\to$ \emph{Calibration} $\to$ \emph{Auction} $\to$ \emph{Execution} $\to$ \emph{Refinement} (see Fig.~\ref{fig:method_overview}). 
Given a user request $x$, the \emph{Planning} module decomposes it into a directed graph of atomic task units. 
In the \emph{Calibration} phase, potential agents estimate their success probabilities using a nested mechanism: a pre-trained calibrator $S$ for base accuracy, and a dynamic post-hoc calibrator $S'$ that adapts to distribution shifts. 
This gives the auction a reliability-adjusted basis for allocation.
The \emph{Auction} module assigns each task unit using a score that balances calibrated confidence against execution cost.
Selected agents \emph{Execute} the subtasks, and a composer synthesizes the \emph{Final Answer}. 
Finally, the \emph{Refinement} module can use evaluator feedback to update $S'$ over time.

\subsection{Task Planning and Unit Extraction}

The \emph{LLM Planner} ($f_{\text{1}}$) first interprets the input $x$ to generate a directed task graph $G=(V,E)$, where nodes represent atomic steps with defined skill requirements and data dependencies (see Appendix~\ref{app:prompts} for detailed prompts).

Subsequently, the \emph{Task-unit Extractor} optimizes this graph by merging tightly coupled nodes into coarser executable \emph{task units} $T$.

Candidate groups are identified via coupling signals---specifically shared I/O, prompt similarity, or explicit control dependencies.

To ensure feasibility, a merge is finalized only if it satisfies a decision gate: (i) the resulting unit must be supported by agent capabilities and (ii) maintain structural validity within the graph.

\subsection{Confidence Calibration}
We define the calibrated confidence $\hat{p}_{ij}$ for agent $a_j$ on task $t_i$ using a hierarchical composition of static and dynamic transformations.

\paragraph{Static Calibration ($S$).} 
To improve calibration across diverse domains, we employ a static calibrator $S$ trained on a \emph{broad-spectrum corpus of diverse public benchmarks} (including math, coding, and multimodal datasets; details in Appendix~\ref{app:calib_details}).
Instead of fitting a single global scaler, $S$ applies a group-specific scaling followed by histogram binning:
\begin{equation}
\hat{p}_0 = S(p_{\text{raw}}) = \mathcal{S}_{\text{bin}}\left( \sigma( \mathbf{w}_g^T \cdot \phi(p_{\text{raw}}) + b_g ) \right),
\end{equation}
where $p_{\text{raw}}$ is the raw confidence, and parameters $(\mathbf{w}_g, b_g)$ are specific to task group $g$, determined by embedding clustering (KD-tree) on the heterogeneous training data.
This allows the system to retrieve appropriate calibration parameters even for unseen task types based on semantic similarity.

\paragraph{Dynamic Calibration ($S'$).}
To handle distribution shifts (e.g., specific scientific workflows not present in the static corpus), we introduce an online dynamic calibrator $S'$.
It refines the static estimate $\hat{p}_0$ via a time-variant transformation:
\begin{equation}
\hat{p}_{\text{final}} = S'(\hat{p}_0; \theta_t) = \sigma\left( \alpha_t \cdot \text{logit}(\hat{p}_0) + \beta_t \right).
\end{equation}
When correctness feedback is available, the parameters $\theta_t = \{\alpha_t, \beta_t\}$ are updated online via gradient descent to minimize the negative log-likelihood of recent auction outcomes.

\subsection{Auction and Execution}

\paragraph{Bid Construction.}
For each task unit $t_i$ and candidate agent $a_j$, we compute a scalar bid $b_{ij}$ based on two components.
First, we apply a concave power-law transformation to the calibrated confidence $\hat{p}_{ij}$:
\begin{equation}
v(\hat{p}_{ij}) = (\hat{p}_{ij})^\gamma, \quad \text{with } \gamma \in (0, 1].
\end{equation}
This transformation compresses the high-confidence regime, making the mechanism less sensitive to minor fluctuations in high-probability estimates (e.g., distinguishing 0.99 from 0.95 becomes less critical than 0.60 from 0.50).
Second, we calculate a \textit{Normalized Cost} $C_{\text{norm}, j} \in [0, 1]$, defined as a weighted sum of monetary expense and latency:
\begin{equation}
\begin{aligned}
C_{\text{norm}, j}
&= w_p \cdot \min\!\left(
    \frac{c^{\text{in}}_j + c^{\text{out}}_j}{C_{\text{ref}}},\; 1
\right) \\
&\quad + w_l \cdot \min\!\left(
    \frac{L_{\text{ref}}}{l_j},\; 1
\right).
\end{aligned}
\label{eq:cost_norm}
\end{equation}
Here, $c^{\text{in}}_j$ and $c^{\text{out}}_j$ represent the agent's unit execution costs (e.g., token prices for LLMs), while $l_j$ denotes its throughput (tokens/sec). These are normalized against a domain-specific reference price ceiling ($C_{\text{ref}}$) and reference throughput ($L_{\text{ref}}$). The weights $w_p$ and $w_l$ govern the trade-off between budget and speed.
\paragraph{Winner Selection.}
The final bid is derived by penalizing the transformed confidence with the weighted cost:
\begin{equation}
b_{ij} = (\hat{p}_{ij})^{\gamma} - \beta \cdot C_{\text{norm}, j}.
\end{equation}
The parameter $\beta \geq 0$ acts as the global \textit{Cost Sensitivity}. A higher $\beta$ biases the system toward cost-efficiency, selecting cheaper or faster agents unless a premium model offers a substantial confidence gain. The winning agent is identified via $j^* = \arg\max_{j} b_{ij}$.

\paragraph{Auction Overhead.}
The auction adds one confidence query per eligible task-unit--agent pair, for $\sum_{t\in T}|A_t|$ bid calls, where $A_t$ is the eligible agent set for unit $t$; this reduces to $|T|\times|A|$ when all agents are eligible.
These independent queries request scalar confidences and can be parallelized. For larger pools, a preliminary top-$k$ filter could reduce bid traffic.

\subsection{Refinement Loop}
To handle distribution shifts in long-horizon deployment, the final stage \emph{optionally} closes the loop by updating the dynamic calibrator $S'$. 
This module is designed to be adaptive: it activates when a feedback mechanism (e.g., unit tests for code or a lightweight verifier) provides a binary correctness label $y_{\text{label}} \in \{0, 1\}$.
We use ground-truth feedback only for upper-bound analysis; without reliable feedback, Agora defaults to the static calibrator $S$.
This label is used to update the parameters $\theta_t$ of $S'$ via gradient descent:
\begin{equation}
\theta_{t+1} \leftarrow \theta_t - \eta \nabla_{\theta} \mathcal{L}_{\text{BCE}}(S'(\hat{p}_0; \theta_t), y_{\text{label}}).
\end{equation}
This update lets $S'$ adapt when correctness feedback is available.

\section{Experiments}
\label{sec:experiments}
\subsection{Datasets}
\noindent\emph{AIME 2025}: We utilize the complete set of 30 problems from the 2025 American Invitational Mathematics Examination to test advanced reasoning, evaluated via Pass@1. Each answer changes accuracy by 3.33 percentage points, so we treat AIME as a small-sample stress test.

\noindent\emph{MMLU-Pro}: To evaluate robustness, we use MMLU-Pro \citep{wang2024mmluprorobustchallengingmultitask}. Due to computational constraints, we evaluate on a stratified random subset of 2,000 examples, reporting the overall accuracy.

\noindent\emph{SciCode}: This benchmark simulates scientific workflows \citep{tian2024scicoderesearchcodingbenchmark}. We report Pass Rates on the full test split. Following official recommendations, we utilize the \texttt{with\_background} setting to strictly test knowledge retrieval.

\noindent\emph{SPIQA}: For multimodal context, we use the Test-A split of SPIQA \citep{pramanick2025spiqadatasetmultimodalquestion}, reporting Retrieval Accuracy and L3 Score.

\noindent\emph{MathVision}: We assess visual mathematical reasoning using the \emph{testmini} split of MathVision \citep{wang2024measuringmultimodalmathematicalreasoning}, reporting Pass@1.

\subsection{Baselines}We compare our proposed framework against a diverse set of baselines, ranging from static single-model executions to advanced routing and cascading strategies. Since not all baselines support multimodal inputs, we categorize them based on their compatibility with text-based and visual inference tasks (see Table~\ref{tab:baselines}).

\begin{table}[t]
    \centering
    \caption{Overview of baseline methods and their compatibility with text-based versus multimodal inference tasks ($\checkmark$: Supported; $\times$: Not supported/evaluated).}
    \label{tab:baselines}
    \resizebox{\columnwidth}{!}{%
    \begin{tabular}{llcc}
        \toprule
        \textbf{Type} & \textbf{Method} & \textbf{Text} & \textbf{Vision} \\
        \midrule
        \multirow{2}{*}{Static} 
          & Single Strong / Weak & \checkmark & \checkmark \\
          & Random Router & \checkmark & \checkmark \\
        \midrule 
        Heuristic 
          & 1NN Router (Text Embed) & \checkmark & $\times$ \\ 
        \midrule 
        \multirow{2}{*}{Cascade} 
          & Consistency Cascade & \checkmark & \checkmark \\
          & FrugalGPT & \checkmark & \checkmark \\          
        \midrule 
        \multirow{2}{*}{Learned} 
          & AdaptiveLLM & \checkmark & $\times$ \\          
          & Hybrid LLM (Best-Route) & \checkmark & $\times$ \\ 
        \bottomrule
    \end{tabular}%
    }
\end{table}

\noindent\emph{Static and Heuristic Baselines}: To establish performance bounds, we evaluate \textit{Single Strong/Weak} executions, which exclusively utilize the most capable or cost-effective model for all queries. We also include a \textit{Random Router} as a stochastic lower bound. For retrieval-based routing, we employ a \textit{1NN Router} that selects agents based on text embedding similarity for language tasks.

\noindent\emph{Cascading Strategies}: To address cost-efficiency, we evaluate sequential frameworks. \textit{Consistency Cascade} \citep{wang2023selfconsistencyimproveschainthought} starts with a smaller model and escalates to a stronger one only if the consistency of sampled outputs falls below a threshold. \textit{FrugalGPT} \citep{chen2023frugalgptuselargelanguage} queries a chain of models ordered by cost, stopping early if the answer is deemed reliable; we adapt this for visual tasks by chaining vision-language models.

\noindent\emph{Learned Routers}: Finally, we compare against representative predictors trained to optimize routing. Our adapted implementation of \textit{AdaptiveLLM} \citep{cheng2025adaptivellm} uses MiniLM query embeddings with a lightweight benchmark-specific classifier. \textit{Hybrid LLM} \citep{ding2024hybridllmcostefficientqualityaware} (Best-Route) explicitly models query difficulty and agent expertise to maximize expected accuracy under specific budget constraints.

\subsection{Implementation}
\label{sec:implementation}

\paragraph{Model Selection.}
To test the framework's adaptability, we employed distinct model pairs for each benchmark. Within each benchmark, all routing and cascade methods use the same candidate pool:\\
(1) \emph{AIME 2025}: \texttt{xai/grok-4-1} (Strong) paired with \texttt{gpt-oss-20b} (Weak);\\
(2) \emph{MMLU-Pro}: \texttt{Mistral-Small-3.2-24B} (Strong) paired with \texttt{qwen3-14b} (Weak);\\
(3) \emph{SciCode}: \texttt{xai/grok-4-1} (Strong) paired with \texttt{openai/gpt-5-mini} (Weak).\\
(4) For multimodal tasks \emph{(SPIQA, MathVision)}, we consistently used \texttt{Qwen/Qwen3-VL-Thinking} and \texttt{xai/grok-4-1-fast-vision}.

\paragraph{System Configuration.}
To demonstrate the controllability of Agora, we define three operating modes based on the cost sensitivity $\beta$:
(1) \emph{Quality-First} ($\beta=0.001$): Prioritizes accuracy, using cost only as a tie-breaker. This is our default setting for matched-pool accuracy comparisons.
(2) \emph{Balanced} ($\beta=0.1$): Seeks a trade-off between performance and budget.
(3) \emph{Cost-Efficient} ($\beta=0.5$): Aggressively optimizes for savings, comparable to frugal baselines.
All operating modes use the same static calibrators; changing $\beta$ only changes the cost penalty and requires no retraining.
In our main results (Table~\ref{tab:text_results}), we report the \emph{Quality-First} performance to compare against strong baselines, while the trade-off characteristics are analyzed in Sec.~\ref{sec:sensitivity}.

\paragraph{Planning Configuration.}
While our framework generally employs an LLM Planner, we adapted this for specific benchmarks to leverage intrinsic structures. For \textit{AIME 2025} and \textit{MMLU-Pro}, the planner decomposes queries into explicit reasoning steps. However, for \textit{SciCode} and \textit{SPIQA}, we bypassed the planner to utilize their pre-defined sub-problem and retrieval-reasoning structures, respectively. This isolation allows us to evaluate the auction mechanism's efficacy purely on task allocation, decoupling it from planning quality.

\paragraph{Baseline Implementation Details.}
To ensure a rigorous comparison, we categorize baselines into \textit{training-free} strategies and \textit{learned} routers. 
For \emph{Learned Baselines} (AdaptiveLLM, Hybrid LLM, 1NN), we strictly isolated training data to avoid leakage: 
(1) For \emph{AIME 2025}, we used historical AIME problems from 1983--2024; 
(2) For \emph{MMLU-Pro}, we utilized the remaining examples excluding our stratified test subset; 
(3) For \emph{SciCode}, we used the official validation split.
For \emph{Cascading Strategies}, we implemented inference-time variants without parameter updates: \textit{Consistency Cascade} generates $k=3$ samples and escalates if agreement is low, while \textit{FrugalGPT} employs an LLM-based verifier (confidence threshold $\ge 0.7$) to substitute the original trained scoring function.
Detailed hyperparameters (e.g., consistency thresholds) are provided in Appendix~\ref{app:baselines}.

\subsection{Results}
\begin{table}[t]
    \centering
    \caption{\emph{Text-based reasoning results.} \emph{Vanilla (Strong/Weak)} denotes single-model execution with the candidates in Sec.~\ref{sec:implementation}. AIME Vanilla scores come from the Artificial Analysis standardized evaluation \protect\citep{artificialanalysis2025leaderboard}. \emph{Agora (S/W)} uses the strong/weak model as planner. Results use the \emph{quality-first setting} ($\beta=0.001$); Sec.~\ref{sec:sensitivity} reports cost trade-offs.}
    \label{tab:text_results}
    \resizebox{\columnwidth}{!}{%
    \begin{tabular}{lcccc}
        \toprule
        \multirow{2}{*}{\textbf{Methods}} & \textbf{AIME} & \textbf{MMLU} & \multicolumn{2}{c}{\textbf{SciCode}} \\
        \cmidrule(lr){4-5}
         & \textbf{2025} & \textbf{Pro} & \textbf{Sub} & \textbf{Main} \\
        \midrule
        Vanilla (Strong)     & 89.5 & 68.1 & 44.2 & 13.6 \\
        Vanilla (Weak)       & 71.7 & 64.4 & 39.2 & 12.4 \\
        Random Router        & 76.7 & 66.3 & 42.3 & 12.6 \\
        Consist. Cascade     & 89.5 & 67.0 & \textbf{45.4} & \textbf{14.1} \\
        1NN Router           & 76.7 & 68.3 & 43.0 & 12.3 \\
        AdaptiveLLM          & 86.7 & \underline{69.3} & 43.2 & 13.6 \\
        Hybrid LLM           & 80.0 & 64.5 & 43.2 & 11.2 \\
        FrugalGPT            & 86.7 & 66.7 & \underline{44.4} & 13.6 \\
        \midrule
        \textbf{Agora (S)} & \textbf{93.3} & \textbf{71.9} & \multirow{2}{*}{\underline{44.4}} & \multirow{2}{*}{ \underline{13.8}} \\
        \textbf{Agora (W)} & \underline{90.0} & 69.1 &  &  \\
        \bottomrule
    \end{tabular}%
    }
    
    \vspace{12pt} 
    
    \caption{\emph{Multimodal Results.} Comparison on SPIQA and MathVision (MV). For SPIQA, \emph{Ret} denotes Retrieval Accuracy; \emph{Avg}, \emph{$\ge$0.6}, and \emph{$\ge$0.8} refer to the GPT-4o-evaluated L3 Reasoning Scores.}
    \label{tab:vision_results}
    \resizebox{\columnwidth}{!}{%
    \begin{tabular}{lccccc}
        \toprule
        \multirow{2}{*}{\textbf{Methods}} & \multicolumn{4}{c}{\textbf{SPIQA}} & \multirow{2}{*}{\shortstack{\textbf{Math}\\\textbf{Vision}}} \\
        \cmidrule(lr){2-5}
        
         & \textbf{Ret} & \textbf{Avg} & \textbf{$\ge$0.6} & \textbf{$\ge$0.8} & \\
        \midrule
        
        Vanilla (Grok)  & \textbf{{85.4}} & 60.5 & \underline{65.6} & 43.4 & 50.3 \\
        Vanilla (Qwen)    & 67.3 & 55.2 & 59.2 & \underline{48.2} & 53.3 \\
        Random Router     & 72.0 & 57.6 & 62.2 & 44.4 & 50.0 \\
        Consist. Cascade  & 72.6 & 55.5 & 60.0 & 45.6 & 51.4 \\
        FrugalGPT         & 75.2 & \underline{61.7} & 64.4 & 45.1 & \underline{54.6} \\
        \midrule
        \textbf{Agora}  & \underline{84.9} & \textbf{65.0} & \textbf{72.4} & \textbf{56.9} & \textbf{55.3} \\
        \bottomrule
    \end{tabular}%
    }
\end{table}
We analyze performance across text-based and multimodal benchmarks (Tables~\ref{tab:text_results} and \ref{tab:vision_results}).

\emph{Text-based Reasoning.}
Agora improves matched-pool performance on logic-intensive tasks. On \emph{AIME 2025}, \textit{Agora (S)} reaches \emph{93.3\%}, versus 89.5\% for the strong single-model baseline. On \emph{MMLU-Pro}, it reaches \emph{71.9\%}, the best result among the evaluated methods.
On \emph{SciCode}, Agora obtains a 44.4\% subproblem pass rate, above the strong single-model baseline (44.2\%), tied with FrugalGPT, and below Consistency Cascade (45.4\%). This three-run mean suggests that distribution shift limits routing gains: the calibrator is trained on competitive programming rather than scientific code.

\emph{Multimodal Reasoning.}

Table~\ref{tab:vision_results} reports results in visual domains.
On \emph{SPIQA}, Agora achieves the best Average L3 score among the evaluated methods (\emph{65.05\%}), exceeding FrugalGPT by over 3 percentage points.
On the strict metric ($\ge$0.8), Agora achieves \emph{56.9\%}, an 8.7-point improvement over the best single model (Qwen: 48.2\%).
This result reflects functional complementarity: Grok is stronger at retrieval, whereas Qwen is stronger at strict reasoning.
Together, the results delimit when routing helps: gains are largest when candidate strengths are separable, whereas domain-misaligned calibration yields competitive rather than dominant performance.
For Agora, SciCode and SPIQA use predefined units, so their outcomes isolate allocation from planner quality.

\section{Discussions}
\label{sec:bibtex}
\subsection{Ablation Study}
\label{sec:ablation}

\begin{figure}[t]
    \centering
    \includegraphics[width=0.85\linewidth]{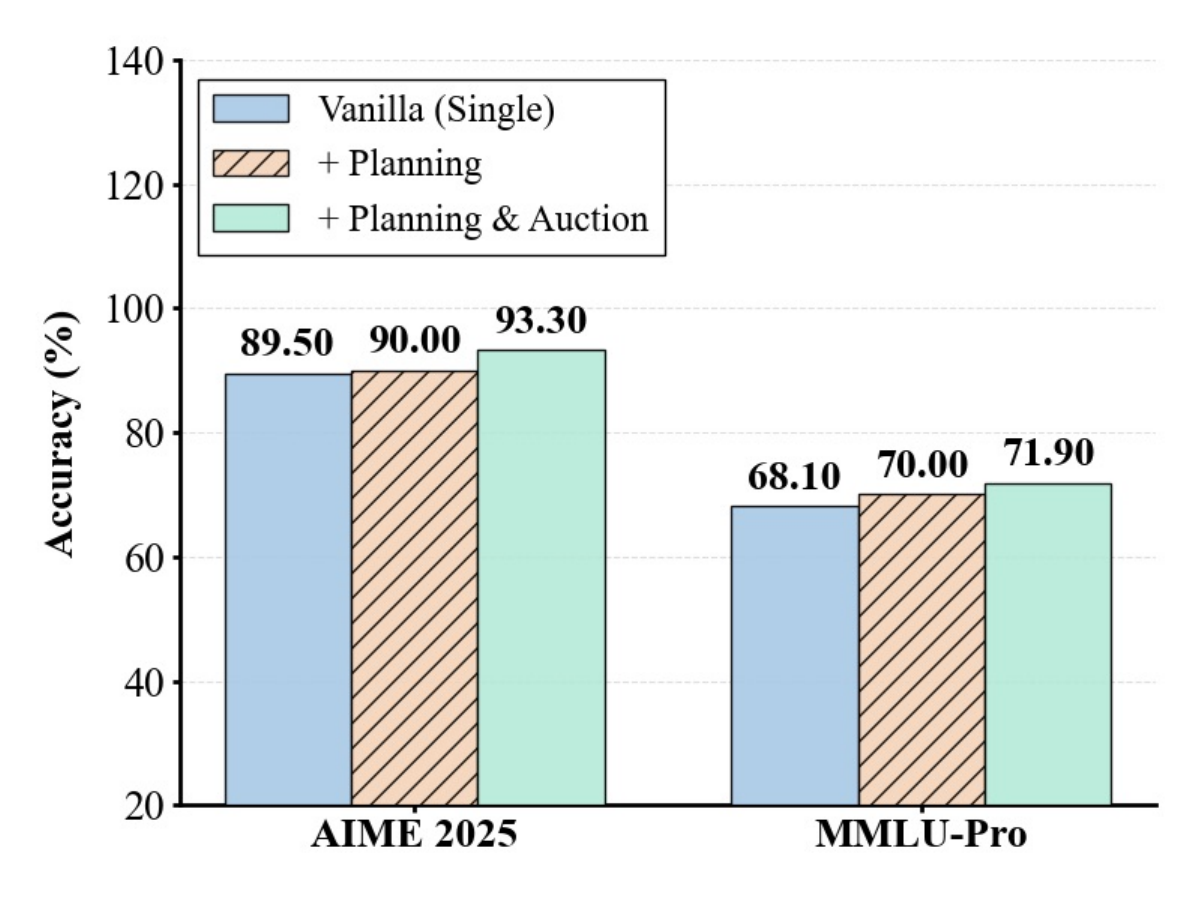}
    \caption{Ablation study on AIME 2025 and MMLU-Pro. The AIME series uses the strong model as planner (\textit{Agora (S)}). Task planning and auction routing yield progressive gains over Vanilla.}
    \label{fig:ablation}
\end{figure}

As shown in Figure~\ref{fig:ablation}, both decomposition and dynamic routing drive consistent gains.
Focusing on MMLU-Pro, which tests robust reasoning across diverse subjects, we observe a step-wise enhancement.
\emph{Task Planning} first lifts the single-model baseline by +1.90\% (from 68.10\% to 70.00\%), demonstrating that structural decomposition reduces problem complexity even for strong generalist models.
Building on this, the \emph{Auction Mechanism} contributes an equal, additional +1.90\% boost, reaching a peak accuracy of \emph{71.90\%}.
This cumulative improvement shows that planning provides the reasoning blueprint, while the auction assigns each sub-task to the highest-scoring agent.

\paragraph{Impact of System Architecture.}
Figure~\ref{fig:ablation} reports the \textit{Agora (S)} AIME chain: 89.50\% $\rightarrow$ 90.00\% $\rightarrow$ 93.30\%.
For \textit{Agora (W)}, task planning raises AIME accuracy from 71.69\% to 83.30\% (11.61 percentage points), and auction routing further raises it to 90.00\% (6.70 points).
The two AIME chains separate these effects: decomposition contributes more with the weak planner, whereas auction routing improves both planner settings.
Given AIME's 30-item size, we treat the exact increments as supporting rather than high-precision component estimates.

\paragraph{Impact of Online Refinement ($S'$).}
The +1.2-point comparison (70.7\% $\rightarrow$ \emph{71.9\%}) uses ground-truth correctness feedback and therefore estimates the potential value of online adaptation, not the default black-box setting.
$S'$ is enabled only when reliable labels are available, such as unit-test or tool-based verification; otherwise Agora retains static $S$.

\subsection{Calibration Efficacy Analysis}
\label{sec:calibration_analysis}

A central premise of our auction mechanism is that an agent's bid must accurately reflect its probability of success. We evaluate this from two perspectives: intrinsic reliability (ECE metrics) and extrinsic impact (downstream accuracy).

\subsubsection{Intrinsic Reliability}
\begin{figure}[t]
    \centering
    \includegraphics[width=\linewidth]{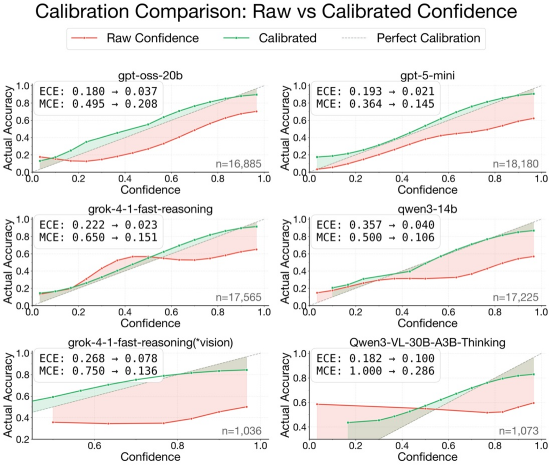}
    \caption{Reliability diagrams comparing raw (red) versus calibrated (green) confidence scores across six representative models. The grey dashed line represents perfect calibration. Metrics shown are Expected Calibration Error (ECE) and Maximum Calibration Error (MCE) before and after calibration.}
    \label{fig:calibration_results}
\end{figure}

Figure~\ref{fig:calibration_results} visualizes the reliability diagrams for six representative agents. 
The curves consistently deviate below the diagonal ($y=x$), indicating that when models predict high confidence (e.g., $>0.8$), their empirical accuracy is substantially lower.
For instance, \texttt{grok-4-1-fast-reasoning} and \texttt{qwen3-14b} show high Expected Calibration Errors (ECE) of 0.222 and 0.357, respectively, rendering their raw logits unreliable for bidding.

Applying our hierarchical calibrator (green curves) improves these probabilities.
ECE decreases across all six models—most notably, \texttt{grok-4-1} improves by an order of magnitude (\emph{0.222 $\to$ 0.023}).
Crucially, this efficacy extends to challenging multimodal domains: the calibrator successfully rectifies the vision-language model \texttt{Qwen3-VL} (ECE 0.182 $\to$ 0.100), enabling the auction to fairly compare visual and textual reasoning bids.

\subsubsection{Extrinsic Impact: Preventing the Winner's Curse}
Calibration is central to reliable bidding. Without it, overconfident agents can win the auction.
Table~\ref{tab:calibration_ablation} measures this effect.

\begin{table}[t]
    \centering
    \caption{Ablation study of calibration impact on downstream accuracy (\%). We compare the Vanilla strong baseline against our Auction mechanism with and without calibration. Values in parentheses denote the absolute performance gain/loss ($\Delta$) relative to the Vanilla baseline. \textit{Note}: For SPIQA, accuracy is reported on the strict subset where L3 Score $\ge$ 0.8.}
    \label{tab:calibration_ablation}
    \resizebox{\columnwidth}{!}{%
    \begin{tabular}{lccccc}
        \toprule
        \multirow{2}{*}{\textbf{Configuration}} & \textbf{AIME} & \textbf{MMLU} & \textbf{SciCode} & \textbf{SPIQA} & \textbf{Math} \\
         & \textbf{2025} & \textbf{Pro} & \textbf{Sub} & \textbf{(L3$\ge$0.8)} & \textbf{Vision} \\
        \midrule
        Vanilla & 89.5 & 68.1 & 44.2 & 48.2 & 53.3 \\
        \midrule
        Auction & 90.0 & 67.5 & 42.3 & 46.9 & 49.3 \\
        \textit{((No Calibration))} & \small{\textcolor{teal}{(+0.5)}} & \small{\textcolor{red}{(-0.6)}} & \small{\textcolor{red}{(-1.9)}} & \small{\textcolor{red}{(-1.3)}} & \small{\textcolor{red}{(-4.0)}} \\
        \midrule
        Auction & \textbf{93.3} & \textbf{71.9} & \textbf{44.4} & \textbf{56.9} & \textbf{55.3} \\
        \textit{(With Calibration)} & \small{\textbf{\textcolor{teal}{(+3.8)}}} & \small{\textbf{\textcolor{teal}{(+3.8)}}} & \small{\textcolor{teal}{(+0.2)}} & \small{\textbf{\textcolor{teal}{(+8.7)}}} & \small{\textbf{\textcolor{teal}{(+2.0)}}} \\
        \bottomrule
    \end{tabular}%
    }
\end{table}

While text-based tasks show minor volatility without calibration, \emph{multimodal tasks suffer a performance collapse}. 
For instance, on MathVision, the uncalibrated auction drops \emph{4.0\%} below the single-model baseline, as the system mistakenly prioritizes agents hallucinating high confidence on incorrect answers.
Incorporating calibration reverses this trend, yielding $\Delta +2.0\%$ on MathVision and $+8.7\%$ on SPIQA.
Appendix~\ref{app:musique} shows the same pattern on MuSiQue. F1 rises from 44.1 for Single Mistral to 51.4 without calibration and 54.3 with calibration.
Across the main suite, the uncalibrated auction falls below the single-model baseline on MMLU-Pro, SciCode, SPIQA, and MathVision.
Auction structure alone is therefore insufficient; calibrated bids are needed for reliable winner selection.
MuSiQue uses dataset-provided decomposition, so its calibration gain is independent of planner quality.

\subsection{Mechanism of Complementarity}
\label{sec:complementarity}

\begin{figure}[t]
    \centering
    \includegraphics[width=\columnwidth]{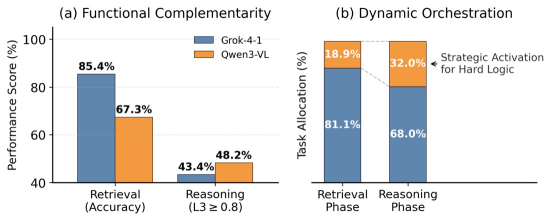}
    \caption{\emph{Mechanism Analysis on SPIQA.} (a) Functional complementarity: Grok excels in retrieval, Qwen in reasoning. (b) Dynamic orchestration: The system strategically shifts logic-intensive sub-tasks to Qwen.}
    \label{fig:mechanism}
\end{figure}

We attribute Agora's gains on SPIQA to the framework's exploitation of \emph{functional orthogonality}. As visualized in Figure~\ref{fig:mechanism}(a), the candidate models possess complementary capability frontiers: \texttt{grok-4-1} acts as a ``Retrieval Specialist'' (85.4\% retrieval accuracy vs. 67.3\% for Qwen), whereas \texttt{Qwen3-VL-Thinking} serves as a ``Reasoning Specialist'' (48.2\% strict reasoning rate vs. 43.4\% for Grok).

Consequently, the auction mechanism orchestrates these distinct strengths (Figure~\ref{fig:mechanism}b). In the \emph{Retrieval phase}, the system routes \emph{81.1\%} of sub-tasks to Grok, shielding the pipeline from Qwen's weaker visual grounding. Conversely, in the \emph{Reasoning phase}, allocation shifts dynamically: while Grok handles easier instances, Qwen is activated for the \emph{32\%} of harder tasks. On these winning bids, Qwen scores 64.2, compared with 53.2 for Grok. This shows that Agora combines complementary strengths rather than defaulting to one model.
Because SPIQA supplies its retrieval--reasoning structure, the stage-specific shift reflects conditional allocation rather than planner quality or a global backend preference.

\subsection{The Cost-Quality Trade-off Analysis}
\label{sec:sensitivity}
\begin{figure}[t]
    \centering
    \includegraphics[width=\linewidth]{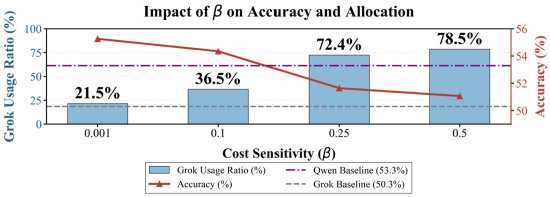}
    \caption{Impact of cost sensitivity $\beta$ on MathVision. Increasing $\beta$ shifts allocation toward the cost-efficient agent (Grok), reducing normalized cost at the expense of marginal accuracy drops.}
    \label{fig:beta_sensitivity}
\end{figure}

To quantify the controllability of our auction mechanism, we conducted a sensitivity analysis on the cost sensitivity parameter $\beta$ using the MathVision benchmark. 
As shown in Figure~\ref{fig:beta_sensitivity}, adjusting $\beta$ acts as a direct lever for governing the trade-off between economic efficiency and reasoning accuracy.

Increasing $\beta$ from $0.001$ to $0.5$ triggers a strategic shift in allocation: the system progressively favors the cost-efficient agent (\texttt{grok-4-1}), with its usage ratio surging from \emph{21.5\% to 78.5\%}.
This aggressive offloading demonstrates the mechanism's responsiveness to cost constraints.
However, this efficiency comes at a price: accuracy decreases from 55.26\% to 51.06\%.
Notably, the transition from $\beta=0.001$ to $\beta=0.1$ reveals a balanced "sweet spot," where Grok usage nearly doubles (to 36.5\%) with only a negligible accuracy drop ($\approx 0.9\%$).
In contrast, higher $\beta$ values ($\ge 0.25$) prioritize budget over performance, pushing accuracy toward the lower bound of the standalone weak model.
Because all $\beta$ settings reuse the same static calibrators, the curve isolates allocation changes induced by the cost penalty rather than gains from retuning.

Bid collection is distinct from executor cost: scalar bid traffic depends on eligible task-unit--agent pairs, whereas $\beta$ controls executor selection.
Parallel queries and preliminary candidate filtering could limit traffic for larger pools.

\section{Conclusion}
We introduced \emph{Agora}, a framework for enhancing LLM reasoning through confidence-calibrated, auction-based task allocation.
By reformulating task allocation as a dynamic marketplace, Agora treats reasoning steps as tradeable items, enabling agents to bid based on verified competence rather than raw confidence.
Our hierarchical calibration reduces the effect of overconfident bids and routes tasks using calibrated competence.
Across five main benchmarks, Agora improves or remains competitive with matched-pool routing baselines, especially when candidate agents have complementary strengths.
It also exposes a controllable cost--quality trade-off without retraining.

\section*{Limitations}

Our framework has three boundaries.

\paragraph{Calibration Generalization.}
Our mechanism relies on calibration quality. Developing a universal calibrator is data-hungry, and decomposed sub-tasks often present out-of-distribution challenges. Under distribution shift, the system can fall back to the strongest agent or require a larger bid margin before rerouting.

\paragraph{Dependency on Planning.}
Decomposition strategies are context-dependent and lack a ``one-size-fits-all'' solution. Tightly coupled tasks may violate the independent auction assumption and can instead be merged into a single task unit. Decomposition can also increase sub-task semantic variance, destabilizing the post-hoc calibrator.

\paragraph{Model Composition.}
Efficiency hinges on agent complementarity. If one model dominates all dimensions, candidate preselection or single-model execution avoids unnecessary bidding.

\bibliography{custom}

\appendix

\section{Prompt Templates}
\label{app:prompts}

To ensure reproducibility, we provide the exact system prompts used in the Agora pipeline. These templates are stored as plain text within the codebase.

\subsection{Planner (Plan Graph Generation)}
The planner is implemented in \texttt{aucteam/planner\_agent/}. The prompt is assembled by concatenating the \texttt{SYSTEM\_PRIMER}, few-shot exemplars, and \texttt{OUTPUT\_INSTRUCTIONS}.

\paragraph{Rationale for Plan Constraints.}
We explicitly instruct the planner to limit output to a maximum of 4 steps. Empirical observation suggests that while longer plans offer finer granularity, they significantly increase the probability of \textit{cascading failures}. The 4-step limit serves as a heuristic regularization.

\paragraph{System Primer.}
\begin{scriptsize}
\begin{verbatim}
You are an expert research planner.
Break complex requests into modular steps
with explicit data/control dependencies.
Represent every plan as a single JSON
object that conforms to this schema:

- Root object:
  * steps (list): ordered execution steps.
  * edges (list): dependencies between IDs.
  * notes (str): clarifications.

- Step object template:
  {"id": "T1", "description": "Outcome",
   "skills": ["capability"],
   "reads": ["art_id"], 
   "writes": ["art_id"]}
  * id: unique string starting with 'T'.
  * description: what to accomplish.
  * skills/reads/writes: always lists.

- Edge object template:
  {"from": "T1", "to": "T2", 
   "type": "data", "why": "Rationale"}
  * type: "data" or "control".

Ensure the JSON is syntactically valid.
\end{verbatim}
\end{scriptsize}

\paragraph{Output Instructions.}
\begin{scriptsize}
\begin{verbatim}
Respond with exactly one JSON object,
followed by <END_JSON> on its own line.
Do not include Markdown fences.

CRITICAL - Plan Requirements:
1. MAXIMUM 4 STEPS.
2. Keep step descriptions HIGH-LEVEL.
3. Do NOT embed:
   - Specific numerical calculations
   - Concrete formulas or equations
   - Intermediate computation results
4. Describe the GOAL, not the METHOD.
\end{verbatim}
\end{scriptsize}

\subsection{Executor (Task-Unit Execution)}
Each task unit is executed with a structured prompt that isolates the unit's context.

\paragraph{Task-Unit Prompt Template.}
\begin{scriptsize}
\begin{verbatim}
Overall task: {task_description}

Executing unit {unit_id}

Steps in this unit:
  - {node_id}: {step_description}
  - ...

Skills: {skills}

=======================================
INPUTS FROM PRIOR STEPS:
=======================================

[{artifact_name}]:
{artifact_content} 
# truncated to 8000 chars per artifact

=======================================

Instructions:
- If final unit:
  - This is the FINAL step.
  - {final_answer_hint_1}
- Else:
  - Focus only on this unit.
  - Provide intermediate findings.
  - Do not give the final answer yet.
  - Use INPUTS FROM PRIOR STEPS.

Context: {optional_context}
\end{verbatim}
\end{scriptsize}

\subsection{Self-Reported Confidence Prompt}
Before the auction, agents self-report a scalar confidence $p_{\mathrm{raw}}$.

\begin{scriptsize}
\begin{verbatim}
Rate your confidence (0-1) that you can
correctly implement this function:

{step_description_prompt} 
# truncated to 800 chars

Function signature:
{function_header} 

Respond with ONLY a number between
0 and 1 (e.g., 0.85).
\end{verbatim}
\end{scriptsize}

\section{Calibration Implementation Details}
\label{app:calib_details}

Our calibration framework is designed to ensure robustness across diverse task types. We explicitly distinguish the training data sources for the static and dynamic components to balance generalization with adaptability.

\paragraph{Static Calibrator ($S$): Broad-Spectrum Training.}
To enable the static calibrator to adapt to unseen task types (e.g., domain-specific scientific problems) and minimize the risk of distribution shift, we trained it on a \emph{comprehensive, large-scale corpus} aggregated from diverse public benchmarks.
Instead of relying on simple binning which fails to capture the complexity of broad-domain data, we utilized the ground-truth labels from these datasets to train the Hierarchical Scaling (HS-QAB) parameters.
Crucially, we prioritized using validation or test splits from source benchmarks where possible to minimize overlap with the pre-training corpora of large language models.

The training corpus is categorized by modality:
\begin{itemize}
    \item \emph{Text-based Reasoning Tasks:} We aggregated datasets covering mathematics, coding, and commonsense reasoning, including:
    \textit{ACEReason (Math)}, \textit{AGIEval}, \textit{AIME (Training split)}, \textit{ARC-Challenge/Easy} (Train/Val/Test), \textit{Codeforces}, \textit{CSQA}, \textit{GSM8K}, \textit{HumanEval}, \textit{MMLU} (Subset), and \textit{TACO}.
    \item \emph{Multimodal Tasks:} We utilized visual reasoning datasets including \textit{MathVerse (Test)}, \textit{SPIQA (Validation)}, \textit{ChatQA}, and \textit{Geometry3K}.
\end{itemize}
This training is intended to provide $S$ with a broad baseline estimate, including for tasks outside the calibration corpus.

\paragraph{Dynamic Calibrator ($S'$): Historical Adaptation.}
While the static calibrator targets broad generalization, the dynamic calibrator $S'$ is trained on the \emph{historical interaction data} accumulated during the current inference session.
This allows the system to correct residual errors and adapt to the specific distribution of the target task in real-time.

\section{Algorithm Details}
\label{app:algo}

\paragraph{Task-Unit Extraction Algorithm.}
Pseudocode for the heuristic rules used to merge tightly coupled nodes.

\begin{scriptsize}
\begin{verbatim}
Algorithm: Task-Unit Extraction

Input: Task Graph G = (V, E)
Params: Bandwidth=2, Jaccard_Thresh=0.6

1. Initialize Candidate Groups S = {}

2. Rule A: Data Reuse Band
   For each pair (u, v) in Bandwidth:
     # Check intersection
     If Intersect(Writes(u), Reads(v)):
       If path u->v is clean:
         Add {u, v} to S

3. Rule B: Prompt Similarity
   For each pair (u, v):
     If Jaccard(u.txt, v.txt) >= Thresh:
       Add {u, v} to S

4. Rule C: Control Gates
   For each control edge u->v:
     Add {u, v} to S

5. Merge & Validate
   Union overlapping groups in S.
   For each Group g in S:
     If Valid(g) AND Capable(g):
       If Savings > Threshold:
         Merge nodes in g
\end{verbatim}
\end{scriptsize}

\section{Additional MuSiQue Evaluation}
\label{app:musique}

We additionally evaluate Agora on MuSiQue-Ans \citep{trivedi2022musique} using the dataset-provided question decompositions as task units. We use 200 training examples for a two-bin calibrator and a disjoint, fixed 500-example development subset for evaluation. The candidate pool contains Mistral-Small-3.2-24B and Qwen3-14B; outputs are generated greedily with seed 9552 and replayed across routing methods. We report Exact Match (EM) and token-level F1. Agora obtains 43.0 EM and 54.3 F1; the F1 95\% bootstrap confidence interval is [50.6, 58.2] over 1,000 resamples.

\begin{table}[ht]
    \centering
    \caption{MuSiQue-Ans results on the fixed 500-example development subset. Agora uses the quality-first setting ($\beta=0.001$).}
    \label{tab:musique_appendix}
    \small
    \begin{tabular}{lcc}
        \toprule
        \textbf{Method} & \textbf{EM} & \textbf{F1} \\
        \midrule
        Single Mistral & 33.6 & 44.1 \\
        Single Qwen & 25.2 & 35.5 \\
        Random Router & 39.2 & 50.4 \\
        1NN Router & \underline{42.4} & \underline{54.1} \\
        Auction w/o Calibration & 39.6 & 51.4 \\
        \textbf{Agora} & \textbf{43.0} & \textbf{54.3} \\
        \bottomrule
    \end{tabular}
\end{table}

\section{Baseline Implementation Details}
\label{app:baselines}

\subsection{Training Data Construction}
For baselines requiring supervision (1NN Router, AdaptiveLLM, Hybrid LLM), we constructed training sets distinct from the evaluation splits:
\begin{itemize}
    \item \emph{AIME 2025}: We used 933 unique historical AIME problems from 1983--2024 for training.
    \item \emph{MMLU-Pro}: The original dataset contains 12k+ examples. We reserved our stratified 2,000-sample test set and used the remaining $\sim$10k examples for training learned routers.
    \item \emph{SciCode}: Following standard protocol, we utilized the provided validation split (featuring distinct problems from the test set) for training.
\end{itemize}

\subsection{Configuration of Cascading Baselines}
\noindent\emph{Consistency Cascade}: This method operates without training. We configured it to first query the cost-efficient model. We generate $k=3$ reasoning paths; if the self-consistency agreement (vote ratio) is below $\tau=0.6$, the system escalates to the strong model.

\noindent\emph{FrugalGPT}: We implement a training-free variant of FrugalGPT \citep{chen2023frugalgptuselargelanguage}: a strong LLM judges the weak model's answer, which is accepted at confidence $\ge 0.7$ and otherwise escalated. This approximates the original learned scorer without a benchmark-specific classifier.

\end{document}